\pdfoutput=1

\documentclass[11pt]{article}

\usepackage{EACL2023}

\usepackage{times}
\usepackage{latexsym}

\usepackage[T1]{fontenc}

\usepackage[utf8]{inputenc}

\usepackage{microtype}

\usepackage{inconsolata}

\usepackage{tabularx}
\usepackage{graphicx}
\usepackage{latexsym}
\usepackage{array,ragged2e}
\newcolumntype{P}[1]{>{\centering\arraybackslash}p{#1}}
\newcolumntype{Y}{>{\centering\arraybackslash}X}
\newcolumntype{L}{>{\raggedright\arraybackslash}X}
\usepackage{todonotes}
\usepackage{diagbox}
\usepackage{colortbl}
\usepackage{csvsimple} 
\usepackage[utf8]{inputenc}
\usepackage{CJK}
\usepackage{hhline}

\title{MEDs for PETs: Multilingual Euphemism Disambiguation for Potentially Euphemistic Terms}
\author{Patrick Lee,   {\bf Alain Chirino Trujillo}, {\bf Diana Cuevas Plancarte}, {\bf Olumide Ebenezer Ojo}, {\bf Xinyi Liu}, \\{\bf Iyanuoluwa Shode}, {\bf Yuan Zhao}, {\bf Jing Peng}, {\bf Anna Feldman} \\
      Montclair State University\\New Jersey, USA\\
 \texttt{\{leep,chirinotruja1,cuevasplancd1,ojoo,liux2,shodei,zhaoy2,pengj,feldmana\}@montclair.edu}}

\begin{document}
\maketitle
\begin{abstract}


This study investigates the computational processing of euphemisms, a universal linguistic phenomenon, across multiple languages. We train a multilingual transformer model (XLM-RoBERTa) to disambiguate potentially euphemistic terms (PETs) in multilingual and cross-lingual settings. In line with current trends, we demonstrate that zero-shot learning across languages takes place. We also show cases where multilingual models perform better on the task compared to monolingual models by a statistically significant margin, indicating that multilingual data presents additional opportunities for models to learn about cross-lingual, computational properties of euphemisms. In a follow-up analysis, we focus on universal euphemistic “categories” such as death and bodily functions among others. We test to see whether cross-lingual data of the same domain is more important than within-language data of other domains to further understand the nature of the cross-lingual transfer.

\end{abstract}

\section{Introduction}

Euphemisms are a linguistic device used to soften or neutralize language that may otherwise be harsh or awkward to state directly (e.g. “between jobs” instead of “unemployed”, “late” instead of “dead”, “collateral damage” instead of “war-related civilian deaths”). By acting as alternative words or phrases, euphemisms are used daily to maintain politeness, mitigate discomfort, or conceal the truth. While they are culturally-dependent, the need to discuss sensitive topics in a non-offensive way is universal, suggesting similarities in the way euphemisms are used across languages and cultures. 
 
This study explores whether multilingual models take advantage of such similarities when processing euphemisms. We use the multilingual transformer model XLM-RoBERTa-base \citep{conneau2020unsupervised}, or “XLM-R”, as our deep learning model, and work with four languages (Mandarin Chinese, American English, Spanish, and {Yor\`{u}b\'{a}}) that encompass a diverse range of linguistic and cultural backgrounds. In our experiments, we focus on the euphemism disambiguation task, in which potentially euphemistic terms (PETs) are classified as euphemistic (1) or not (0) in a given context (e.g., “let go” may mean “fired” in some contexts, but not all in other contexts). Models are trained on labeled data from a single, or multiple languages, and evaluated separately on all four languages.

Our contributions are as follows: (1) We augment existing Chinese and Spanish datasets started by \citet{lee2023feed} and perform additional analyses (Section 3). (2) We run classification experiments and find cases of cross-lingual transfer (i.e. a model trained on one language can classify instances in another language), as well as an overall performance improvement when training models on multiple languages versus one (Section 4). (3) We perform a follow-up experiment in which we find signs that the cross-lingual transfer may be related to euphemistic category (Section 5). These results suggest that XLM-R picks up on “knowledge” about euphemisms which it can not only transfer, but also synergize across languages.

\begin{table*}[!h]
\begin{center}
\scalebox{0.95}{
\begin{tabularx}{\textwidth}{|Y|Y|Y|Y|Y|Y|Y|Y|} 
 \hline
 \textbf{Lang} & \textbf{TotalEx} & \textbf{EuphEx} & \textbf{NonEuphEx} & \textbf{TotPETs} & \textbf{AmbPETs} & \textbf{$\alpha$}\\
 \hline
 EN & 1952 & 1383 & 569 & 129 & 58 & 0.415\\
 \hline
 ZH & 2005 & 1484 & 521 & 110 & 36 & 0.635\\
 \hline
 ES & 1861 & 1143 & 718 & 147 & 91 & 0.576\\
 \hline
 {YO} & 1942 & 1281 & 661 & 129 & 62 & 0.679\\ 
 \hline
\end{tabularx}
}
\end{center}
\caption{Statistics of multilingual datasets used for the euphemism disambiguation experiments.}
\label{tbl:multilingual_stats}
\end{table*}

\section{Related Work}

In recent years, there has been growing interest in computational approaches to euphemism detection in the natural language processing (NLP) community. \citet{felt2020recognizing} introduced the recognition of euphemisms and dysphemisms using NLP, generating near-synonym phrases for sensitive topics.  \citet{zhu2021self} proposed euphemism detection and identification tasks using masked language modeling with BERT. \citet{gavidia-etal-2022-cats} created a corpus of potentially euphemistic terms (PETs). \citet{lee2022searching} developed a linguistically driven approach for identifying PETs using distributional similarities.  BERT-based systems that participated in a shared task on euphemism disambiguation showed promise \cite{lee2022sharedtask}. \citet{keh2022exploring} experimented with classifying PETs unseen during training. \citet{lee2023feed} perform transformer-based euphemism disambiguation experiments, exploring vagueness as one of the properties of euphemisms. 

Other existing work has explored the multilingual and cross-lingual transfer capabilities of large language models (LLMs). \citet{choenni2023languages} found that multilingual LLMs rely on data from multiple languages to a large extent, learning both complementary and reinforcing information. \citet{shode2023nollysenti} found cases where transfer learning from out-of-language data in a particular domain performed better than same-language data in a different domain. 

\section{Multilingual Corpus of Euphemisms}

For our data, we use the multilingual Mandarin Chinese (ZH), American English (EN), Spanish (ES), and {Yor\`{u}b\'{a}} (YO) euphemism datasets created by \citet{lee2023feed}. In these datasets, text examples containing PETs are annotated by native speakers with a 0 or a 1 (i.e. a euphemistic or non-euphemistic usage of the PET). We modify the datasets to become similar to one another in two ways: Firstly, {Yor\`{u}b\'{a}} lacked “boundary tokens” to the left and right side of PETs, so we add them in where possible; for some examples ($\sim$25\%), the PET tokens were sometimes separated due to {Yor\`{u}b\'{a}} word order, so multiple pairs of “boundary tokens” were added for these examples. Secondly, to balance the number of examples in each language, we augmented the Mandarin Chinese and Spanish datasets. Using the guidelines from the original paper, native speakers (who were co-authors) added more PETs (40 for Chinese and 67 for Spanish) and examples (453 for Chinese and 900 for Spanish) to obtain the final euphemism corpus used for this paper\footnote{\url{https://github.com/pl464/euph-detection-datasets/tree/main/EACL\_2024}}. See Table \ref{tbl:multilingual_stats} for the updated metrics. 

As can be seen, while the number of examples are fairly balanced across languages, there are still two main differences. One is the number of ambiguous PETs; i.e. PETs which have both euphemistic and non-euphemistic usages in the dataset. Higher numbers of ambiguous PETs and examples may contribute to a higher “degree of difficulty“ for classification. Two, we additionally contribute interrater agreement metrics for the Mandarin Chinese, Spanish, and {Yor\`{u}b\'{a}} datasets. We recruited 2 native speakers to annotate a random subset of 500 examples from each dataset and then compute Krippendorf's alpha \cite{hayes2007answering}, $\alpha$, following the example of \cite{gavidia-etal-2022-cats} who obtained an alpha of 0.415 for the English dataset. The results can be found in the last column Table \ref{tbl:multilingual_stats}. We believe these two differences may correlate with the “degree of difficulty” in classifying each dataset.

\section{Multilingual and Cross-lingual Experiments}

\subsection{Methodology}
\label{sec:methodology}

For our experiments, we use XLM-R-base, a multilingual transformer model pre-trained on multiple languages, including Mandarin (ZH), English (EN), and Spanish (ES), but not {Yor\`{u}b\'{a}} (YO) \cite{conneau2020unsupervised}. We experiment with fine-tuning XLM-R on euphemism data from multiple languages (when multiple languages are present in the training data, we refer to this as “multilingual”) versus one (“monolingual”). For each test run, we randomly sample 1800 examples from each language and use a 80-10-10 split to create training, validation, and test sets. We create the multilingual train/val sets by combining and shuffling the train/val data from multiple languages (e.g., the training set for the 4-language setting consists of 5760 examples– 1440 of each language). The test sets are held constant across all settings so that we can observe the impact of including multiple languages during training. 



Our non-default fine-tuning parameters were: batch size=16, learning rate=1e-5, max epochs=30, and early stopping patience=5. We performed 30 test runs for each training setting (e.g. ZH, ES+EN, etc), each time using the best trained model (before early stopping) for inference on the test set; using 4 NVIDIA Tesla A100 GPUs, fine-tuning 30 times took approximately 6 hours for each language present in the training set.

\subsection{Results}

The results of these experiments are in Table \ref{tbl:results_1}. The values shown are averaged Macro-F1 scores across the 30 runs\footnote{Standard deviations generally ranged from 0.02-0.04.}. Note that for each cell in the table, the row shows the training language(s) (“All” refers to training on all four languages), while the column shows the test language. For example, the average Macro-F1 score when training on Chinese data but testing on English data was 0.653. A majority-class baseline is provided. Additionally, the colored cells indicate cases where the language of the test set appeared in the training set.

Firstly, as expected, the performances of the monolingual models tested on the same language (green cells) are significantly better than the baseline. We noted the unusually high performance of Chinese (0.895), which was also the dataset with the smallest range of PETs. So, we followed up by repeating the monolingual fine-tuning experiments, but restricting the data in each language to cover exactly 52 PETs spanning 815 examples. The results, shown in Appendix \ref{sec:appendix_a}, show much more balanced results, suggesting that performance is impacted by the range of PETs present in the data. 

Secondly, we observed an extent of zero-shot, cross-lingual learning taking place with the monolingual models (white cells). For instance, the English-on-Chinese score was 0.607, and Spanish-on-English was 0.639. In general, there appeared to be similar interactions between Chinese, English, and Spanish, with scores ranging from 0.535-0.653. By comparison, the monolingual models performed poorly on {Yor\`{u}b\'{a}}, with scores ranging from 0.300-0.384. The monolingual {Yor\`{u}b\'{a}} models, too, did not perform very well on the other languages, although not as poorly (0.383-0.417). This suggests something transferable between Chinese, English, and Spanish, but not as much for {Yor\`{u}b\'{a}}, possibly due to language-specific factors (i.e. {Yor\`{u}b\'{a}} euphemisms differ significantly from the others) or the fact that XLM-R was not pre-trained on {Yor\`{u}b\'{a}} data. Interestingly, we observed slightly higher cross-lingual scores when replicating the experiments at a smaller number of examples (1500), the results of which are shown in Appendix \ref{sec:appendix_b}. Further testing is needed to investigate the relationship between data size and cross-lingual performance.

Lastly, we observed that the performances of the multilingual models were generally higher than those of the monolingual models. The boldfaced values in each column indicate the best setting for that test language, which was always multilingual. We observe more specific trends in the “bilingual” (blue) and “trilingual” (purple) results: for Chinese, the English data contributes the most, and vice versa; Spanish benefits from all other languages, but more so Chinese and English; {Yor\`{u}b\'{a}} mostly benefits from English. For each test language, we assess the statistical significance between the best (boldfaced) multilingual scores and the monolingual scores by computing the paired t-test value (p=0.05) across the 30 test runs. The resulting t-test values are as follows: Chinese, 0.0011; English, 6e-7; Spanish, 0.0047; {Yor\`{u}b\'{a}}, 0.074. From this, we conclude that the effect of including data from all 4 languages was statistically significant for Chinese, English and Spanish, but not {Yor\`{u}b\'{a}}. Further, the varying “contributions” across different language combinations suggests that specific language relationships come into play when performing multilingual euphemism disambiguation.

\begin{table}[!h]
\begin{center}
\scalebox{1.0}{
\begin{tabular}{|l|c|c|c|c|} 
 \hline
 \diagbox[height=2.5em]{\textbf{Train}}{\textbf{Test}} & \textbf{ZH} & \textbf{EN} & \textbf{ES} & \textbf{YO} \\
 \hline
 \textbf{Baseline} & \cellcolor{gray!15}0.426 & \cellcolor{gray!15}0.416 & \cellcolor{gray!15}0.381 & \cellcolor{gray!15}0.394 \\
 \hline
 \textbf{ZH} & \cellcolor{green!25}0.879 & 0.653 & 0.535 & 0.300 \\
 \hline
 \textbf{EN} & 0.607 & \cellcolor{green!25}0.765 & 0.567 & 0.381 \\
 \hline
 \textbf{ES} & 0.613 & 0.639 & \cellcolor{green!25}0.752 & 0.384 \\
 \hline
 \textbf{YO} & 0.417 & 0.407 & 0.383 & \cellcolor{green!25}0.790 \\
 \hline
 \textbf{ZH+EN} & \cellcolor{cyan!15}0.897 & \cellcolor{cyan!15}0.804 & 0.508 & 0.397 \\
 \hline
 \textbf{EN+ES} & 0.650 & \cellcolor{cyan!15}0.781 & \cellcolor{cyan!15}0.764 & 0.416 \\
 \hline
 \textbf{ES+YO} & 0.605 & 0.630 & \cellcolor{cyan!15}0.758 & \cellcolor{cyan!15}0.794 \\
 \hline
 \textbf{ZH+ES} & \cellcolor{cyan!15}0.884 & 0.670 & \cellcolor{cyan!15}0.764 & 0.377 \\
 \hline
 \textbf{EN+YO} & 0.616 & \cellcolor{cyan!15}0.772 & 0.602 & \textbf{\cellcolor{cyan!15}0.802} \\
 \hline
 \textbf{ZH+YO} & \cellcolor{cyan!15}0.881 & 0.646 & 0.585 & \cellcolor{cyan!15}0.795 \\
 \hline
 \textbf{ZH+EN+ES} & \cellcolor{blue!12}0.898 & \textbf{\cellcolor{blue!12}0.805} & \cellcolor{blue!12}0.775 & 0.389 \\
 \hline
 \textbf{EN+ES+YO} & 0.647 & \cellcolor{blue!12}0.783 & \cellcolor{blue!12}0.772 & \cellcolor{blue!12}0.791 \\
 \hline
 \textbf{ZH+EN+YO} & \cellcolor{blue!12}\textbf{0.899} & \cellcolor{blue!12}0.801 & 0.555 & \cellcolor{blue!12}0.794 \\
 \hline
 \textbf{ZH+ES+YO} & \cellcolor{blue!12}0.885 & 0.664 & \textbf{\cellcolor{blue!12}0.778} & \cellcolor{blue!12}0.778 \\
 \hline
 \textbf{All} & \cellcolor{blue!30}0.895 & \cellcolor{blue!30}0.792 & \cellcolor{blue!30}0.776 & \cellcolor{blue!30}0.793 \\
 \hline
\end{tabular}
}
\caption{Average Macro-F1s for the multilingual and cross-lingual experiments}
\label{tbl:results_1}
\end{center}
\end{table}

\section{Experiments with Euphemistic Category}

Motivated by the question “what is the nature of the cross-lingual knowledge being learned about euphemisms?”, we ran a follow-up experiment in which we looked at specific euphemistic categories\footnote{All PETs were assigned categories in the datasets.}. We created test sets of examples in which we isolate a single language and a single category, out of a possible 4 categories that had a substantial number of examples in each dataset: physical/mental attributes (ATTR), bodily functions/parts (BODY), death (DEATH), and sexual activity (SEX). Then, we compare two different training settings: (1) training only on same-category, but out-of-language examples (“SC-OOL”), and (2) training only on same-language, but out-of-category examples (“SL-OOC”). For all language-category scenarios, there were always fewer SC-OOL examples than SL-OOC, so we used the maximum number of SC-OOL examples available, down-sampled for the SL-OOC examples, and used a random 90-10 split to create training and validation sets. More detailed metrics regarding the number of examples can be found in Appendix \ref{sec:appendix_c}. We use the same parameters as in \ref{sec:methodology}, except we increased the early stopping patience to 10 (due to having smaller datasets) and only perform 10 runs for each setting.

In Table \ref{tbl:topic_results}, we show the differences in average Macro-F1 scores between the SC-OOL and SL-OOC settings. That is, positive values (green) indicate that the SC-OOL setting performed better, whereas negative values (red) indicate the opposite; e.g. for the test set containing Chinese ATTR euphemisms, training on English, Spanish, and {Yor\`{u}b\'{a}} ATTR euphemisms yielded an average F1 of 0.088 points higher than when training on Chinese euphemisms from other categories. 
We observed that SC-OOL examples performed better than SL-OOC in 7 out of the 16 language-category scenarios. While this is interesting, since we would expect that training on same-language examples should generally perform better, there are no obvious patterns with either language or category (except perhaps that Spanish did not generally benefit from SC-OOL examples). Despite this, the results suggest the overall possibility that examples which contribute cross-lingual understanding are related by semantic category. More testing, particularly with specific language combinations and categories, may reveal more definitive cross-lingual results. Additionally, the full tables of Macro-F1 scores for each setting (which can be found in Appendix \ref{sec:appendix_d}) show that the overall scores were low. This indicates the overall challenge of classifying examples with PETs not seen during training, even to the extent that out-of-language examples could outperform within-language examples. 
\begin{table}[!h]
\begin{center}
\begin{tabularx}{\columnwidth}{|l|Y|Y|Y|Y|} 
 \hline
 \textbf{Lang} & \textbf{ATTR} & \textbf{BODY} & \textbf{DEATH} & \textbf{SEX} \\
 \hline
 \textbf{ZH} & \cellcolor{green!25}+0.088 & \cellcolor{green!25}+0.083 & \cellcolor{red!10}-0.026 & \cellcolor{red!25}-0.094 \\
 \hline
 \textbf{EN} & \cellcolor{red!10}-0.038 & \cellcolor{green!10}+0.034 & \cellcolor{red!25}-0.288 & \cellcolor{green!25}+0.069 \\
 \hline
 \textbf{ES} & \cellcolor{red!10}-0.007 & \cellcolor{red!48}-0.303 & \cellcolor{red!10}-0.019 & \cellcolor{red!25}-0.097 \\
 \hline
 \textbf{YO} & \cellcolor{green!38}+0.12 & \cellcolor{green!10}+0.042 & \cellcolor{green!10}+0.011 & \cellcolor{red!25}-0.094 \\
 \hline
\end{tabularx}
\end{center}
\caption{Differences in Macro-F1 scores on category-specific test sets between the “SC-OOL” and “SL-OOC” settings.}
\label{tbl:topic_results}
\end{table}

\section{Conclusions and Future Work}

In this study, we investigate the multilingual and cross-lingual capabilities of multilingual transformers for euphemism disambiguation. We found cases of zero-shot, cross-lingual learning, and that fine-tuning on multiple languages yields statistically significant improvements for Chinese, English, and Spanish. This indicates that multilingual approaches may work as a method of data augmentation, which would be particularly useful for data-scarce figurative language tasks (especially for low-resource languages). The results also suggest that some of these patterns are language-specific, and dependent on training settings. More work is needed to test other training parameters (e.g. number of examples) and languages from a variety of families. 

While it is hard to answer the question “what exactly is being learned about euphemisms cross-lingually?”, we found preliminary evidence that part of the answer may relate to euphemisms' semantic category. Exploring this question further is left to future work, which may be important from both a linguistic and computational perspective. 


\section*{Limitations}

While the terms “Chinese” and “English” were sometimes used for brevity, the Chinese data used in this study only included Mandarin data, while the English data only includes American English. (However, the Spanish and {Yor\`{u}b\'{a}} data are from a variety of dialects.) Additionally, XLM-R is taken to be representative of other transformer/multilingual deep learning models, and the impact of XLM-R's pre-training scheme was not investigated. We did not conduct a thorough search for hyperparameters (which were selected mostly based on prior work), and limited computational resources prevented experimentation with other (larger) multilingual language models, such as XLM-R-large.

\section*{Ethics Statement}
The authors foresee no ethical concerns with the work presented in this paper.

\section*{Acknowledgements}
This material is based upon work supported by the
National Science Foundation under the Grant number 2226006.

\bibliography{anthology,custom,euphs}



\bibliographystyle{acl_natbib}

\newpage
\onecolumn
\appendix

\section{Experiments Balanced for PETs}
\bigskip
\label{sec:appendix_a}
The results below show the monolingual models' performances when the number of unique PETs in the sampled data for each setting was held constant (52 PETs spanning 815 examples). Fine-tuning parameters were the same, except for early stopping patience, which was set to 8 (instead of 5) due to the smaller datasets sometimes needing more epochs to converge. 30 runs were still performed for each setting. As can be seen, the performance of the monolingual Chinese (ZH) model on the Chinese test sets is now more similar to the others, though there are still differences between languages which were seen in the main experiments (e.g. Spanish-on-Spanish performance being the lowest; Chinese and {Yor\`{u}b\'{a}} being the highest).

\bigskip
\begin{table}[!h]
\begin{center}
\begin{tabular}{|l|c|c|c|c|} 
 \hline
 \diagbox[height=2.5em]{\textbf{Train}}{\textbf{Test}} & \textbf{ZH} & \textbf{EN} & \textbf{ES} & \textbf{YO} \\
 \hline
 \textbf{ZH} & \cellcolor{green!25}0.749 & 0.594 & 0.611 & 0.363 \\
 \hline
 \textbf{EN} & 0.548 & \cellcolor{green!25}0.727 & 0.589 & 0.370 \\
 \hline
 \textbf{ES} & 0.561 & 0.615 & \cellcolor{green!25}0.710 & 0.445 \\
 \hline
 \textbf{YO} & 0.365 & 0.353 & 0.358 & \cellcolor{green!25}0.752 \\
 \hline
\end{tabular}
\caption{Average Macro-F1s for the monolingual models when examples are constrained to the same number of PETs in the data}
\label{tbl:appendix-a}
\end{center}
\end{table}

\section{Experiments with a Smaller Number of Examples (1500)}
\bigskip
\label{sec:appendix_b}
The results below show the monolingual models' performances when a fewer number of examples were used for train-val-test splits than the main experiments (1500 vs. 1800). Fine-tuning parameters were the same, and 30 runs were performed for each setting. While the monolingual models' performances on the same languages (green cells) were generally lower, some of the zero-shot, cross-lingual performances (white cells) were higher than those in Table \ref{tbl:results_1}.

\bigskip
\begin{table}[!h]
\begin{center}
\begin{tabular}{|l|c|c|c|c|} 
 \hline
 \diagbox[height=2.5em]{\textbf{Train}}{\textbf{Test}} & \textbf{ZH} & \textbf{EN} & \textbf{ES} & \textbf{YO} \\
 \hline
 \textbf{ZH} & \cellcolor{green!25}0.847 & 0.664 & 0.571 & 0.338 \\
 \hline
 \textbf{EN} & 0.615 & \cellcolor{green!25}0.756 & 0.609 & 0.420 \\
 \hline
 \textbf{ES} & 0.600 & 0.628 & \cellcolor{green!25}0.716 & 0.398 \\
 \hline
 \textbf{YO} & 0.411 & 0.417 & 0.401 & \cellcolor{green!25}0.767 \\
 \hline
\end{tabular}
\caption{Average Macro-F1s for the monolingual models using 1500 examples per test}
\label{tbl:appendix-b}
\end{center}
\end{table}


\newpage
\section{Numbers of Examples in the Euphemistic Category Experiments}
\bigskip
\label{sec:appendix_c}
The tables below show the number of examples used in the test sets for each language/category setting in the follow-up study on euphemistic categories. 

\begin{table*}[!h]
\begin{center}
\scalebox{0.9}{
\begin{tabularx}{\textwidth}{|Y|Y|Y|Y|Y|} 
 \hline
 \textbf{Lang} & \textbf{ATTR} & \textbf{BODY} & \textbf{DEATH} & \textbf{SEX}\\
 \hline
 \textbf{ZH} & 157 & 324 & 451 & 501 \\
 \hline
 \textbf{EN} & 573 & 83 & 348 & 89 \\
 \hline
 \textbf{SP} & 311 & 258 & 105 & 111 \\
 \hline
 \textbf{YO} & 151 & 584 & 459 & 637 \\
 \hline
\end{tabularx}
}
\end{center}
\caption{Metrics for the Euphemistic Category Experiment Test Sets}
\label{tbl:appendix_c-test}
\end{table*}

\bigskip
The tables below show the number of examples sampled for the training and validation sets for each language/category setting. 

\begin{table*}[!h]
\begin{center}
\scalebox{0.9}{
\begin{tabularx}{\textwidth}{|Y|Y|Y|Y|Y|} 
 \hline
 \textbf{Lang} & \textbf{ATTR} & \textbf{BODY} & \textbf{DEATH} & \textbf{SEX}\\
 \hline
 \textbf{ZH} & 1035 & 925 & 912 & 837 \\
 \hline
 \textbf{EN} & 619 & 1166 & 1015 & 1249 \\
 \hline
 \textbf{ES} & 881 & 991 & 1258 & 1227 \\
 \hline
 \textbf{YO} & 1041 & 665 & 904 & 701 \\
 \hline
\end{tabularx}
}
\end{center}
\caption{Metrics for Euphemistic Category Experiments Train/Val Sets}
\label{tbl:appendix_c-train}
\end{table*}

\section{Actual Performances of the SC-OOL and SL-OOC Tests from the Euphemistic Category Experiments}
\bigskip
\label{sec:appendix_d}

The averaged F1s for each language/category scenario using the SC-OOL training sets are shown below.

\bigskip
\noindent

\begin{table}[!h]
\begin{center}
\scalebox{0.9}{
\begin{tabularx}{\textwidth}{|Y|Y|Y|Y|Y|} 
 \hline
 \textbf{Lang} & \textbf{ATTR} & \textbf{BODY} & \textbf{DEATH} & \textbf{SEX}\\
 \hline
 \textbf{ZH} & 0.598 & 0.588 & 0.564 & 0.420 \\
 \hline
 \textbf{EN} & 0.602 & 0.438 & 0.556 & 0.650 \\
 \hline
 \textbf{ES} & 0.541 & 0.431 & 0.458 & 0.495 \\
 \hline
 \textbf{YO} & 0.489 & 0.560 & 0.432 & 0.484 \\
 \hline
\end{tabularx}
}
\end{center}
\caption{Average Macro-F1 Scores for the “SC-OOL” experiments}
\label{tbl:appendix_d-SCOOL}
\end{table}

The averaged F1s for each language/category scenario using the SL-OOC training sets are shown below.

\bigskip
\noindent

\begin{table}[!h]
\begin{center}
\scalebox{0.9}{
\begin{tabularx}{\textwidth}{|Y|Y|Y|Y|Y|} 
 \hline
 \textbf{Lang} & \textbf{ATTR} & \textbf{BODY} & \textbf{DEATH} & \textbf{SEX}\\
 \hline
 \textbf{ZH} & 0.510 & 0.505 & 0.591 & 0.515 \\
 \hline
 \textbf{EN} & 0.640 & 0.404 & 0.650 & 0.582 \\
 \hline
 \textbf{ES} & 0.548 & 0.733 & 0.477 & 0.592 \\
 \hline
 \textbf{YO} & 0.367 & 0.518 & 0.421 & 0.578 \\
 \hline
\end{tabularx}
}
\end{center}
\caption{Average Macro-F1 Scores for the “SL-OOC” experiments}
\label{tbl:appendix_d-SLOOC}
\end{table}


\bigskip
\noindent

\end{document}